\documentclass[letterpaper, 10 pt, journal, twoside]{IEEEtran}
\ifCLASSINFOpdf
\else
\fi

\usepackage{amsmath,amsfonts}
\usepackage{algorithm}
\usepackage{mathtools}
\usepackage{array}
\usepackage{textcomp}
\usepackage{stfloats}
\usepackage{url}
\usepackage[normalem]{ulem}
\usepackage{verbatim}
\usepackage{graphicx}
\usepackage{subfigure}
\usepackage{caption}
\usepackage{algpseudocode}
\usepackage{tablefootnote}
\usepackage{threeparttable}
\usepackage{cite}
\usepackage{hyperref}
\hypersetup{colorlinks=true, linkcolor=blue, anchorcolor=blue, citecolor=blue}
\usepackage[utf8]{inputenc} 
\usepackage[T1]{fontenc}    
\usepackage{url}            
\usepackage{booktabs}       
\usepackage{nicefrac}       
\usepackage{microtype}      
\usepackage{longfbox}
\usepackage{amssymb}
\usepackage{amsmath}   
\usepackage{amsthm}
\usepackage{bbm}
\usepackage{wrapfig}
\usepackage{multirow}
\usepackage{makecell}
\usepackage{parnotes}
\usepackage{cancel}
\usepackage{float}
\usepackage{chngcntr}
\usepackage{apptools}
\usepackage{enumitem}
\usepackage{framed,enumitem} 
\usepackage[most]{tcolorbox}
\usepackage[algo2e,algoruled,boxed,lined,norelsize]{algorithm2e} 

\graphicspath{{figures/}}

\DeclareMathOperator{\logsumexp}{{\text{LogSumExp}}}
\DeclareMathOperator{\lse}{\texttt{\textbf{LSE}}}

\DeclareMathOperator{\csdf}{\texttt{\textbf{C-SDF}}}
\DeclareMathOperator{\dsdf}{\texttt{\textbf{D-SDF}}}

\newcommand{\norm}[1]{\left\lVert#1\right\rVert}

\newcommand*{\tran}{^{\mkern-1.5mu\mathsf{T}}}

\begin{document}

\title{ ContactSDF: Signed Distance Functions as Multi-Contact Models for Dexterous Manipulation}

\makeatletter
\g@addto@macro\@maketitle{
\setcounter{figure}{0}
  \vspace{-20pt}
  \begin{figure}[H]
  \setlength{\linewidth}{\textwidth}
  \setlength{\hsize}{\textwidth}
  \centering
  \includegraphics[height=5.2cm]{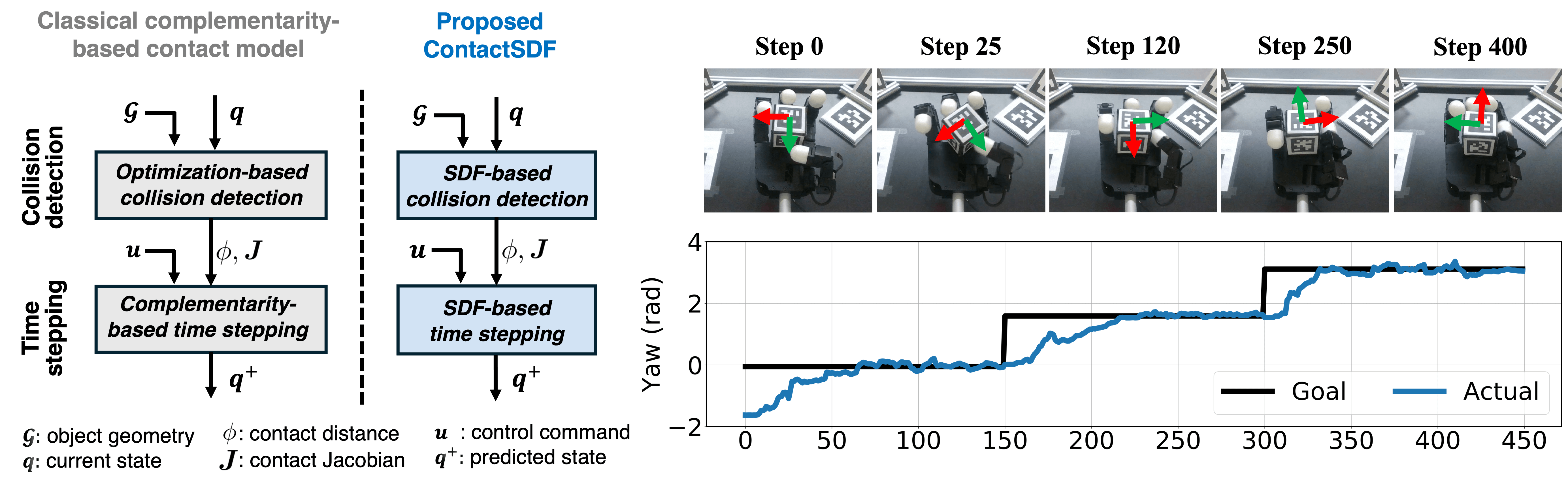}
  \caption{\small{ Left: classic multi-contact model vs. proposed ContactSDF. ContactSDF approximates the contact detection and time-stepping routines using signed distance functions, leading to an explicit and differentiable contact model, which facilitates model-based planning and learning. 
  Right: ContactSDF model predictive control (ContactSDF-MPC) for   Allegro hand on-palm reorientation. Within  2 minutes of learning on hardware, ContactSDF-MPC enables the Allegro hand to perform continuous in-hand reorientation at control frequency of 50 Hz.}}
  \label{quadruped_hardware}
  \end{figure}
  \vspace{-35pt}
}

\author{Wen Yang$^{1}$, Wanxin Jin$^{1}$
\thanks{$^{1}$Both the Authors are with the School for Engineering of Matter, Transport and Energy, Arizona State University, Tempe, AZ 85282 USA.
        {\tt\footnotesize wenyang, wanxin.jin@asu.edu}}
}

\markboth{IEEE Robotics and Automation Letters. Preprint Version.}
{Yang \MakeLowercase{\textit{et al.}}: ContactSDF:Signed Distance Functions as
Multi-Contact Models for Dexterous Manipulation} 

%



\maketitle

\begin{abstract}
In this paper, we propose ContactSDF, a method that uses signed distance functions (SDFs) to approximate multi-contact models, including both collision detection and time-stepping routines. ContactSDF first establishes an SDF using the supporting plane representation of an object for collision detection, and then uses the generated contact dual cones to build a second SDF for time-stepping prediction of the next state. Those two SDFs create a differentiable and closed-form multi-contact dynamic model for state prediction, enabling efficient model learning and optimization for contact-rich manipulation.  We perform extensive simulation experiments to show the effectiveness of ContactSDF for model learning and real-time control of dexterous manipulation. We further evaluate the ContactSDF on a hardware Allegro hand for on-palm reorientation tasks. Results show with around 2 minutes of learning on hardware, the ContactSDF achieves high-quality dexterous manipulation at a frequency of 30-60Hz. \href{https://yangwen-1102.github.io/contactsdf.github.io}{Project page}.
\end{abstract}


%
\IEEEpeerreviewmaketitle

\vspace{-20pt}
\section{Introduction}
\vspace{-5pt}
\IEEEPARstart{I}{n} dexterous manipulation, a robot must decide when and where to make and break contact with objects. The contact-rich interactions make the system dynamics hybrid and non-smooth, creating significant challenges in model learning and control.
While model-free reinforcement learning has shown impressive results in dexterous manipulation tasks, it comes at the cost of extensive data. Conversely, model-based methods, though theoretically more data-efficient, often fall short of achieving similar performance. Recent advances in contact model learning \cite{jinlearninglcs,pfrommer2021contactnets}  and planning \cite{fast_mpc,dairlabc3tro,pang2023global, moura2022non} highlight the importance of handling the hybrid structures of contact dynamics. The complementarity-based formulations \cite{stewart2000implicit,anitescu1997formulating} are widely used in those methods. While effective in representing contact interactions,  complementarity-based models are computationally non-smooth and hybrid, making them difficult to integrate into standard learning and planning frameworks without specialized algorithm design \cite{parmar2021fundamental,dairlabc3tro, marcucci2020warm}. {
Recent studies \cite{fast_mpc, pang2023global} employ interior-point methods to smooth contact dynamics. However, it remains challenging to use the resulting implicit dynamics for real-time optimization-based planning with a trade-off between differentiablity and accuracy.}

To address the above computation challenges, we introduce ContactSDF, a  smoothed multi-contact contact model utilizing SDFs. {ContactSDF has two components to handle discontinuity as shown in Fig. \ref{quadruped_hardware}: (1) SDF-based collision detection, where robot and environment points query a \emph{geometric SDF} constructed directly from the object mesh or point cloud \cite{CVXnet};  and (2) SDF-based time-stepping, where a \emph{velocity-space SDF} is proposed to approximate the state prediction of the multi-contact dynamics based on the contact detection results.  ContactSDF achieves a closed-form forward prediction and end-to-end smoothing (collision detection included). This makes ContactSDF easily integrated into model learning and optimization frameworks, allowing for approximate yet efficient representation for learning and real-time control in dexterous manipulation.} Our contributions are summarized as follows:

\vspace{-5pt}
\begin{itemize}[leftmargin=*]
    \item We propose ContactSDF, a multi-contact model built upon SDFs featuring closed-form prediction and differentiability.
    
    \item We develop ContactSDF predictive control (ContactSDF-MPC), to achieve real-time control of dexterous manipulation. ContactSDF-MPC is then integrated into an on-MPC learning framework to further improve its performance. 

    \item We evaluate the ContactSDF in simulation and on a hardware
Allegro hand for in-hand dexterous manipulation (Fig. \ref{quadruped_hardware}). The results show within 2 minutes of learning on hardware, ContactSDF achieves high-quality control performance for dexterous manipulation with MPC frequency of 50 Hz.
\end{itemize}

\section{Related Works}\label{sec.related_works}
\vspace{-5pt}

\subsection{Complementarity-Based Contact Models}

The rigid body multi-contact dynamics is classically formulated as complementarity-based models \cite{stewart2000implicit, anitescu1997formulating, pang2018robust}, which describes non-smooth constraints on body motions and contact force: the contact force can only arise when two bodies are in contact and must vanish when bodies separate. The complementarity constraints can also be used to describe Coulomb friction, leading to nonlinear complementarity problems, which are difficult to solve. In \cite{anitescu2006optimization}, Anitescu proposed a relaxation to the complementarity problem by changing the non-penetration constraints into a quadratic dual cone constraint. While this relaxation introduces some mild non-physical artifacts \cite{pang2021convex}, it obtains better computational properties and can be considered as the KKT optimality condition of a convex optimization problem \cite{anitescu2006optimization}. 
{The proposed ContactSDF in this paper is derived from the optimization-based model. It is a SDF representation to approximate the optimization-based contact dynamics. Because of approximation only in the primal space (velocity stepping), ContactSDF may not adhere strictly to physical accuracy or dual space interpretation, but it is sufficient for our primary objectives of model-based control and planning.}



\vspace{-15pt}
\subsection{Model-based method for Dexterous Manipulation}
The multi-modality and non-smoothness of complementary-based models make them difficult to use in planning and control for manipulation. For systems with few contact modes, \cite{aceituno2020global, marcucci2020warm} formulate contact planning as mixed-integer programs (MIPs). For contact-rich systems, \cite{dairlabc3tro} proposes decoupling the planning horizon into small MIPs and using the alternating direction method of multipliers for parallel computation. \cite{taskdrivenjin} proposes reduced hybrid models that only capture the contact modes necessary for given tasks. {Another strategy for contact-rich planning, developed by \cite{fast_mpc, pang2023global},   is to smooth  contact dynamics using the log-barrier penalty functions or complementarity relaxation. However, the resulting relaxed contact dynamics is still implicit, requiring a carefully designed optimization process for time stepping. In contrary, ContactSDF
 is a SDF in velocity space, which, by definition of SDF, directly gives the ‘minimal’ distance for a queried unconstrained velocity, and  the one-step gradient projection directly
outputs the contact-constrained velocity. Thus, no optimization process needed for time stepping. {Also, compared to the log-barrier smoothing which only handles contact-resolution (contact model switching)  discontinuity, ContactSDF smooths the entire pipeline of contact modeling, including both collision detection (geometry) and contact-resolution discontinuity. This makes ContactSDF more computationally friendly to optimization-based contact-rich planning and control.}

\vspace{-15pt}
\subsection{Reinforcement Learning for Dexterous Manipulation}
Reinforcement learning (RL) methods have achieved impressive results in dexterous manipulation \cite{qi2022hora,chen_sr_visual_dexterity}. Model-free RL has the benefit of requiring less domain knowledge,  but the cost is they require  millions or billions of environment data \cite{allshire2022transferring}, making them difficult to be applied in hardware. Model-based RL \cite{nagabandi2020deep} shows better data efficiency, but unstructured models may struggle to represent the  multi-modality of contact interaction \cite{pfrommer2021contactnets,parmar2021fundamental}, leading to lower performance compared to model-free approaches. The proposed method generally belongs to the model-based category. Differently,  the proposed ContactSDF model is a physical-based representation derived from classic multi-contact models. Due to the physics inductive bias, our method requires only minutes of training data to accomplish a contact-rich manipulation task in hardware.

\vspace{-15pt}
\subsection{Signed Distance Function Representation}
Signed distance function (SDF) \cite{curless1996sdf} has been widely used in computer vision for geometry learning and reconstructions \cite{dai2017bundlefusion, CVXnet, wen2023bundlesdf}.
 The robotics community has started exploring its use in robot planning and manipulation \cite{Driess2021LearningMA, li2024configurationspacedistancefields, weng2023ngdf, nerualjsdfral, suresh2023neural}. For example, \cite{Driess2021LearningMA} proposes utilizing neural SDF as a state transition model to facilitate manipulation. \cite{weng2023ngdf} formulates a neural grasp distance field as SDF to indicate the task completion distance. \cite{suresh2023neural} proposes jointly learning a neural SDF and estimating object pose using vision and tactile information for in-hand manipulation.
Our  ContactSDF is also inspired by those works. But different from unstructured SDF models learned from data, ContactSDF is physics-based, constructed from the multi-contact dynamics. Hence, the SDFs in ContactSDF have direct physical correspondences: in collision detection, the SDF represents object geometry, and in time stepping prediction, the SDF approximates a projection to the feasible contact velocities satisfying contact constraints.

\vspace{-15pt}
\section{Preliminary and Problem Statement}\label{sec.preliminaries}

\subsection{Quasi-dynamic Optimization-based Contact Model}\label{section:Convex Complementarity Quasi-Dynamic Contact Model}

We use quasi-dynamic models \cite{mason2001mechanics} to describe the equation of motion of a manipulation system state $\boldsymbol{q}:=[\boldsymbol{q}_r, \boldsymbol{q}_o]$, including an actuated robot's position $\boldsymbol{q}_r\in\mathbb{R}^{n_r}$, and unactuated object's position $\boldsymbol{q}_o\in\mathbb{R}^{n_o+1}$.  Quasi-dynamic models have the benefits of simplicity by ignoring the inertial and Coriolis forces that are less significant in slow dexterous manipulation \cite{aceituno2020global,pang2023global,jin2024complementarity}.
Formally, consider a manipulation system with $n_c$ potential contacts, following \cite{pang2021convex,anitescu2006optimization},  the quasi-dynamic  model can be written as a quadratic program (QP):
\vspace{-5pt}
\begin{equation}
\begin{aligned}
    \min_{\boldsymbol{v}} \quad &\frac{1}{2}h^2\boldsymbol{v}\tran\boldsymbol{Q}\boldsymbol{v}-h\boldsymbol{b}\tran \boldsymbol{v}\\
    \text{s.t.} \quad & 
 \boldsymbol{J}_{i,j}\boldsymbol{v}+\frac{\phi_i}{h}\geq 0, i \in\{1...n_c\},\,\,\,j \in\{1...n_d\},
\label{equ:qs_dyn_simplified}
\end{aligned}
\end{equation}
\begin{equation}\label{equ.q_matrix}
\text{with }\quad 
\boldsymbol{Q}:=\begin{bmatrix}
     \boldsymbol{M}_o/h^2 & \boldsymbol{0}\\
    \boldsymbol{0}& \boldsymbol{K}_r
\end{bmatrix},\,\,
    \boldsymbol{b}(\boldsymbol{u}):=\begin{bmatrix}
        m_o\boldsymbol{g}\\
        \boldsymbol{K}_r \boldsymbol{u}+\boldsymbol{\tau}_r
    \end{bmatrix},
\end{equation}
\begin{equation}
    \boldsymbol{J}_{i,j}:=\boldsymbol{J}_{i}^n-\mu_i\boldsymbol{J}_{i,j}^{d}.
\end{equation}
Here, $h$ is the time step.  $\boldsymbol{v}:= [\boldsymbol{v}_o, \boldsymbol{v}_r]$ is the system velocity stacking the object velocity $\boldsymbol{v}_o\in\mathbb{R}^{n_o}$ and robot velocity $\boldsymbol{v}_r\in\mathbb{R}^{n_r}$. $\boldsymbol{M}_o\in\mathbb{R}^{n_o\times n_o}$ is the regularized  mass matrix of the object \cite{pang2023global}, and $m_o\boldsymbol{g}$ is the gravity force of the object with mass $m_o$ and gravity constant $\boldsymbol{g}$. The  robot is assumed in impedance control and treated as a `spring-like' model \cite{hogan1984impedance} with the stiffness matrix  $\boldsymbol{K}_r \in \mathbb{R}^{n_r \times n_r}$. $\boldsymbol{\tau}_r\in\mathbb{R}^{n_r}$ is the non-contact force (e.g., gravity) applied to robot.  The robot input 
$\boldsymbol{u}\in \mathbb{R}^{n_r}$ is the desired position displacement.
{At contact $i=1,2,...,n_c$, 
 the  Coulomb
frictional cone is approximated using a polyhedral cone \cite{stewardtrinklePolyhedron}, with   unit vectors $\left\{\boldsymbol{d}_{i,1}, \boldsymbol{d}_{i,2}, ...,\boldsymbol{d}_{i,n_d}\right\}$ to symmetrically span the contact tangential plane  \cite{stewart2000implicit}.} $\phi_i$ is the signed distance at contact $i$. $\boldsymbol{J}^n_i$ is the system Jacobian to the contact normal, $\boldsymbol{J}^d_{i,j}$ is the system Jacobian to the unit tangential directional vector $\boldsymbol{d}_j$. $\mu_i$ is the friction coefficient. {Due to the page limit, we direct the reader to \cite{pang2021convex,anitescu2006optimization,jin2024complementarity} for the detailed derivation of (\ref{equ:qs_dyn_simplified}).}

In time-stepping prediction,  $\boldsymbol{J}_{i,j}$ is calculated at the system's current state $\boldsymbol{q}$ from a collision detection routine.
The solution $\boldsymbol{v}^{+}$ to (\ref{equ:qs_dyn_simplified}) will be used to integrate from $\boldsymbol{q}$ to the next $\boldsymbol{q}^{+}$, i.e., 
$\boldsymbol{q}^{+}:=\boldsymbol{q}``+"h\boldsymbol{v}^+$,
where ``+"  involves quaternion integration.

\vspace{-15pt}
\subsection{Model Predictive Control}\label{sec.review_mpc}
\vspace{-5pt}
Model predictive control (MPC) can be formulated as
\begin{equation}\label{equ.generic_mpc}
\begin{aligned}
\min_{\boldsymbol{u}_{0:T-1}} \quad 
&  
\sum\nolimits_{t=0}^{T-1} \textit{c}_t\left ( \boldsymbol{q}_t, \boldsymbol{u}_t \right ) +\textit{c}_T\left (\boldsymbol{q}_T\right) 
\\ \text{s.t.}\quad &
\boldsymbol{q}_{t+1} = \boldsymbol{f} \left ( \boldsymbol{q}_{t}, \boldsymbol{u}_{t} \right ), \ t = 0,...,T{-}1, \quad \text{given} \ \boldsymbol{q}_0
\end{aligned}
\end{equation}
where   dynamic model  $\boldsymbol{f} \left ( \cdot \right )$ predicts  future states of a system. Given initial $\boldsymbol{q}_0$,   (\ref{equ.generic_mpc}) is solved for the optimal input sequence $\boldsymbol{u}^*_{0:T-1}(\boldsymbol{x}_0)$ over a prediction horizon $T$, by minimizing the path  $\textit{c}_t(\cdot)$ and final cost $\textit{c}_T(\cdot)$.
For a real system, the MPC controller is implemented in a receding horizon. Specifically, at rollout step $k$, the real system state  is $\boldsymbol{q}_k^{\text{real}}$, and the MPC controller sets 
$\boldsymbol{q}_0=\boldsymbol{q}_k^{\text{real}}$ and solves (\ref{equ.generic_mpc}). Only the first optimal input $\boldsymbol{u}_0^*(\boldsymbol{q}_k^{\text{real}})$ is applied to the real system,  evolving the real system to the next state $\boldsymbol{q}^{\text{real}}_{k+1}$. This receding horizon process repeats and creates a closed-loop control effect, i.e., feedback from real system state $\boldsymbol{q}^{\text{real}}_{k}$ to control input $\boldsymbol{u}^*_0$.

\vspace{-15pt}
\subsection{Problem Statement}
\vspace{-5pt}
In  MPC for contact-rich manipulation, at rollout step $k$ with system state $\boldsymbol{q}_k^{\text{real}}$,  there are two   routines inside MPC:
\begin{itemize}[leftmargin=*]
    \item \emph{Collision detection routine}, which is to identify all potential contacts at  $\boldsymbol{q}^{\text{real}}_{k}$. This routine returns contact distance $\phi_i$ and contact Jacobians $\boldsymbol{J}_{i,j}$ required in (\ref{equ:qs_dyn_simplified}).
    \item \emph{Time-stepping prediction routine}, which is to predict the future system states using the QP model (\ref{equ:qs_dyn_simplified}) given  $\boldsymbol{q}_0=\boldsymbol{q}_k^{\text{real}}$.
\end{itemize}

While both routines can be solved by existing tools, such as mature collision detection algorithms \cite{gilbert1988fast}, and quadratic programming solvers \cite{stellato2020osqp}. However, each routine will create an optimization layer inside the MPC optimization (\ref{equ.generic_mpc}), i.e., $\boldsymbol{f}$ in (\ref{equ.generic_mpc}) includes two optimization layers. As a result, (\ref{equ.generic_mpc}) is a multi-level optimization problem,  creating challenges in computation and achieving real-time control.

In the following paper, we will show that each of the above routines can be replaced using a separate SDF, and thus system model $\boldsymbol{f}$, which includes  collision detection and time-stepping routines, becomes explicit and differentiable. This will eliminate the needs of solving multi-layers of optimizations. Consequently, contact-rich MPC (\ref{equ.generic_mpc})  can be solved efficiently with any regular trajectory optimization or MPC solver \cite{Andersson2019casadi}.

\begin{figure*}[ht]
  \centering
  \includegraphics[scale=0.50]{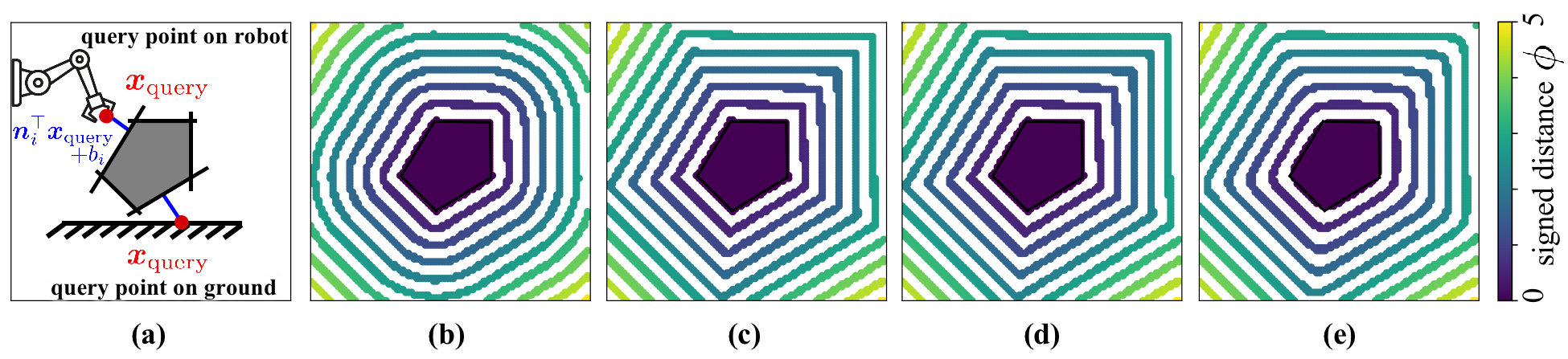}
  \caption{\small 2D  contact detection example. (a) Collision detection in a manipulation system can be viewed as a distance query to a convex object, where query points are taken from the robot and environment. (b) The truth distance field $\phi_{\mathcal{G}}(\boldsymbol{x}_{\text{query}})$ in (\ref{equ.phi}) for the object. (c) Distance field approximated by (\ref{equ.closest_dist}). (d) The $\csdf_{\mathcal{G}}$ distance field  (\ref{equ.lse_dist}) with large $\sigma$. (e)  The $\csdf_{\mathcal{G}}$ distance field (\ref{equ.lse_dist}) with small $\sigma$.  }
  \label{fig:qp_geometric}
  \vspace{-15pt}
\end{figure*}

\vspace{-15pt}
\section{ContactSDF Model}\label{section:Differentiable Truncated Signed Distance Function for a Polytope}
We propose ContactSDF, which includes a $\csdf$  to approximate the collision detection and a $\dsdf$ to approximate the time-stepping prediction in contact-rich manipulation systems.

\vspace{-15pt}
\subsection{\texorpdfstring{$\csdf$}{}: Collision Detection SDF}\label{sec.sdf_for_collision_detection}
\vspace{-5pt}
We assume the geometry of an object in a manipulation system is given and convex, as required in many collision detection algorithms \cite{gilbert1988fast,mamou2009simple}. 
The object geometry can be represented by a set of supporting planes \cite{CVXnet}
\begin{equation}\label{equ.dist_convexset}
    \mathcal{G} =\left\{\boldsymbol{x}\,|\,\boldsymbol{n}_i\tran\boldsymbol{x}+ 
b_i
\leq0,\,\,i=1,2,...,I\right\},
\end{equation}
 each   parameterized by its unit normal vector $\boldsymbol{n}_i \in \mathbb{R}^{3}    $ and offset $b_i \in \mathbb{R}$. The supporting plane representation for object geometry can  be directly obtained from depth data integration \cite{dai2017bundlefusion}, or learned from raw sensor data \cite{CVXnet}.

For collision detection, we take an object-centric perspective. Collision detection is to find the distance $\phi_{\mathcal{G}}(\boldsymbol{x}_{\text{query}})$ between a query point $\boldsymbol{x}_{\text{query}}\in\mathbb{R}^3$ and a closest point on the object $\mathcal{G}$:
\begin{equation}\label{equ.phi}
\begin{aligned}
\phi_{\mathcal{G}}(\boldsymbol{x}_{\text{query}}) :=\min_{\boldsymbol{x}\in\mathcal{G}} \norm{\boldsymbol{x}-
\boldsymbol{x}_{\text{query}}}_2
\end{aligned}.
\end{equation}
Here, a query point $\boldsymbol{x}_{\text{query}}$ can be taken from the surfaces of a robot or environment (e.g., ground).
 The distance becomes zero when the query point is inside the object (contact or penetration). 
$\phi_{\mathcal{G}}(\boldsymbol{x}_{\text{query}})$ in (\ref{equ.phi}) can be considered as a truncated SDF for the object geometry representation $\mathcal{G}$. Getting $\phi_{\mathcal{G}}(\boldsymbol{x}_{\text{query}})$ requires solving an optimization. In the following, we will provide an SDF for fast approximation of (\ref{equ.phi}), inspired by \cite{CVXnet}.

To motivate, we give a  2D example in Fig. \ref{fig:qp_geometric}. In Fig. \ref{fig:qp_geometric}(a), given a point $\boldsymbol{x}_{\text{query}}$ (on robot or ground) and an object $\mathcal{G}$,  the true contact distance field $\phi(\boldsymbol{x}_{\text{query}})$ is shown in Fig. \ref{fig:qp_geometric}(b). One can note that the signed distance between the $\boldsymbol{x}_{\text{query}}$ to each plane $(\boldsymbol{n}_i,b_i)$ can be calculated as  $\boldsymbol{n}\tran_i\boldsymbol{x}_{\text{query}}+b_i$. Thus, with large number $I$ of supporting planes in $\mathcal{G}$,
 we can approximate $\phi_{\mathcal{G}}(\boldsymbol{x}_{\text{query}})$ in  (\ref{equ.phi}) using  $\max$ function,
\begin{equation}\label{equ.closest_dist}\phi_{\mathcal{G}}(\boldsymbol{x}_{\text{query}})\approx
\max\big(0, \max_{i=1,2,...,I}\{\boldsymbol{n}\tran_i\boldsymbol{x}_{\text{query}}+b_i\}
\big),
\end{equation}
as  in Fig. \ref{fig:qp_geometric}(c). To facilitate differentiability of (\ref{equ.closest_dist}), we apply $\logsumexp$  ($\lse$)  function to replace the inner and outer $\max$, as used in \cite{CVXnet}. This yields
\begin{equation}\label{equ.lse_dist}
\begin{aligned}
\small
    \phi_{\mathcal{G}}(\boldsymbol{x}_{\text{query}})\approx
\underbrace{\frac{1}{\sigma}\lse
\bigg\{
0, \lse\big\{\sigma(\boldsymbol{n}_i\tran \boldsymbol{x}_{\text{query}}{+}b_i), i=1,...,I\big\}
\bigg\}}_{:=\csdf_{\mathcal{G}}(\boldsymbol{x}_{\text{query}})}.
\end{aligned}
\end{equation}
Here,  $\lse(\cdot)$ operator is defined as 
\begin{equation}\label{equ.lse}
\lse\{z_1, z_2, ...., z_n\}=\log\sum\nolimits_{i=1}^I \exp(z_i).
\end{equation}

The hyperparameter $\sigma$ in (\ref{equ.lse_dist}) controls the approximation accuracy:  larger $\sigma$ means (\ref{equ.lse_dist}) is closer to (\ref{equ.closest_dist}). Fig. \ref{fig:qp_geometric} (d), (e) shows the distance fields of $\csdf_{\mathcal{G}}(\boldsymbol{x}_{\text{query}})$ with different  $\sigma$.

With  $\csdf_{\mathcal{G}}(\boldsymbol{x}_{\text{query}})$ in (\ref{equ.lse_dist}) approximating $\phi_{\mathcal{G}}(\boldsymbol{x}_{\text{query}})$ in (\ref{equ.phi}), the $\boldsymbol{x}_{\text{query}}$'s closest point $\boldsymbol{x}^*$ on object $\mathcal{G}$, which is the solution to (\ref{equ.phi}),  can be retrived by
\begin{equation}\label{equ.first_approx_2d_tsdf}
\begin{gathered}
    \boldsymbol{x}^*(\boldsymbol{x}_{\text{query}}) {=} \boldsymbol{x}_{\text{query}}
    {-}\csdf_{\mathcal{G}}(\boldsymbol{x}_{\text{query}}) \nabla\csdf_{\mathcal{G}}(\boldsymbol{x}_{\text{query}}).
\end{gathered}
\end{equation}
Intuitively, $\nabla\csdf_{\mathcal{G}}(\boldsymbol{x})$ is an approximated unit directional vector to the nearest point $\boldsymbol{x}^*$ on $\mathcal{G}$, and $\csdf_{\mathcal{G}}(\boldsymbol{x}_{\text{query}})$ is the corresponding distance to $\boldsymbol{x}^*$. (\ref{equ.first_approx_2d_tsdf}) provides an explicit differentiable alternative to collision detection (\ref{equ.phi}). With the above approximated contact distance $\phi(\boldsymbol{x}_{\text{query}})$ and its on-object closest point $\boldsymbol{x}^*(\boldsymbol{x}_{\text{query}})$, contact Jacobian  $\boldsymbol{J}_{i,j}$ required in (\ref{equ:qs_dyn_simplified}) can be computed using differential kinematics \cite{spong2020robot}.

\vspace{-15pt}
\subsection{\texorpdfstring{$\dsdf$}{}: Time-Stepping Prediction SDF}\label{section:Complementarity-Free Explicit Contact Model}

We next show the optimization-based time-stepping model (\ref{equ:qs_dyn_simplified}) can also be approximated by an SDF. {First, we equivalently reformulate the time-stepping model in (\ref{equ:qs_dyn_simplified}) by re-scale and add a constant term $\boldsymbol{b}\tran\boldsymbol{Q}^{-1}\boldsymbol{b}$, which lead the following}
\begin{equation}\label{equ.qs_model2}
\begin{aligned}
    \min_{\boldsymbol{v}} \quad &||{h\boldsymbol{Q}^{\frac{1}{2}}\boldsymbol{v}-\boldsymbol{Q}^{-\frac{1}{2}}\boldsymbol{b}(\boldsymbol{u)}}||_2\\
    \text{s.t.} \quad & \frac{\phi_i}{h}+ \boldsymbol{J}_{i,j}(\boldsymbol{q})\boldsymbol{v}\geq0,
    \,\,i \in\{1...n_c\},\,\,j \in\{1...n_d\}.
\end{aligned}
\end{equation}
Here contact Jacobian $\boldsymbol{J}_{i,j}$ and distance $\phi_i$ is obtained in  $\csdf$ collision detection at system state $\boldsymbol{q}$. If one defines
\begin{equation}\label{equ.def_z}
    \begin{aligned}
\text{new variable}\quad &\boldsymbol{z}:=h\boldsymbol{Q}^{\frac{1}{2}}\boldsymbol{v},\\
\text{and new query point}\quad  &\boldsymbol{z}_{\text{query}}:=\boldsymbol{Q}^{-\frac{1}{2}}\boldsymbol{b}(\boldsymbol{u}),
\end{aligned}
\end{equation}
then the linear constraints in (\ref{equ.qs_model2}) can be re-written  as 
\begin{equation}\label{equ.polytope}
        \mathcal{C}(\boldsymbol{q}):=\biggl\{\boldsymbol{z}\,|\,\boldsymbol{n}_{i,j}\tran\boldsymbol{z} +{b}_{i,j}
    \leq0,\,\,
    \substack{i \in\{1...n_c\}\\ j \in\{1...n_d\}}\biggr\},
\end{equation}
\begin{equation}
\text{with} \quad \small
        \boldsymbol{n}_{i,j}:= -
    \frac{\boldsymbol{Q}^{-\frac{1}{2}}
    \boldsymbol{J}_{i,j}\tran}{||\boldsymbol{Q}^{-\frac{1}{2}}
    \boldsymbol{J}_{i,j}\tran||},
    \quad
    b_{i,j}:=-\frac{ \phi_{i}}{{||\boldsymbol{Q}^{-\frac{1}{2}}
    \boldsymbol{J}_{i,j}\tran||}}. \quad
\end{equation}
Since $\mathcal{C}(\boldsymbol{q})$ is state-dependent,  we here explicitly write its dependence. With the above change of variable, the optimization-based time-stepping model (\ref{equ.qs_model2}) equivalently becomes 
\begin{equation}\label{equ.qs_model3}
    \phi_{\mathcal{C}(\boldsymbol{q})}(\boldsymbol{z}_{\text{query}})=\min_{\boldsymbol{z}\in\mathcal{C}(\boldsymbol{q})} \quad \norm{\boldsymbol{z}-\boldsymbol{z}_{\text{query}}}_2.
\end{equation}
Equation\emph{(\ref{equ.qs_model3}) can be thought of as a collision detection problem in the   $\boldsymbol{z}$-space (image of $\boldsymbol{v}$ space) given  $\boldsymbol{z}_{\text{query}}=\boldsymbol{Q}^{-\frac{1}{2}}\boldsymbol{b}(\boldsymbol{u})$}. Thus, similar to (\ref{equ.lse_dist}),  one can define
\begin{multline}\label{equ.lse_dist2}
     \dsdf_{\mathcal{C}(\boldsymbol{q})}(\boldsymbol{z}_{\text{query}}):=
     \frac{1}{\sigma}\lse
\Bigl(
0, 
\lse
\\
\bigl\{
\sigma(\boldsymbol{n}_{i,j}\tran\boldsymbol{z}_{\text{query}}+b_{i,j}), i \in\{1...n_c\}, j \in\{1...n_d\}
\bigr\}
\Bigr)
\end{multline}
to approximate $\phi_{\mathcal{C}(\boldsymbol{q})}(\boldsymbol{z}_{\text{query}})$ in (\ref{equ.qs_model3}). Further,  one can approximate the optimal solution $\boldsymbol{z}^+$ to (\ref{equ.qs_model3}) using
\begin{equation}\label{equ.qs_model6}
    \boldsymbol{z}^+ \text{=} \boldsymbol{z}_{\text{query}}
    \text{-} \dsdf_{\mathcal{C}(\boldsymbol{q})}(\boldsymbol{z}_{\text{query}}) \nabla\dsdf_{\mathcal{C}(\boldsymbol{q})}(\boldsymbol{z}_{\text{query}}).
\end{equation}
Note that  $\boldsymbol{z}^+$  is in the $\boldsymbol{z}$-space. Recall  definition (\ref{equ.def_z}), we map $\boldsymbol{z}^+$ back to the $\boldsymbol{v}$-space by $ \boldsymbol{v}^+ = \frac{1}{h}\boldsymbol{Q}^{-\frac{1}{2}}\boldsymbol{z}^+.$

In sum, with (\ref{equ.def_z}) and (\ref{equ.qs_model6}) one can finally write the explicit $\dsdf$ time-stepping model, directly mapping from the current state $\boldsymbol{q}$ and  input $\boldsymbol{u}$ to the next velocity $\boldsymbol{v}^+$, as
\begin{multline}\label{equ.vnext}\small
    \boldsymbol{v}^+ = \frac{1}{h} \boldsymbol{Q}^{-1}\boldsymbol{b}(\boldsymbol{u})
    - \frac{1}{h}\boldsymbol{Q}^{-\frac{1}{2}}\biggl(\dsdf_{\mathcal{C}(\boldsymbol{q})}\left(\boldsymbol{Q}^{-\frac{1}{2}}\boldsymbol{b}(\boldsymbol{u})\right)\\
    \nabla\dsdf_{\mathcal{C}(\boldsymbol{q})}\left(\boldsymbol{Q}^{-\frac{1}{2}}\boldsymbol{b}(\boldsymbol{u})\right)\biggr).
\end{multline}
The next state $\boldsymbol{q}^+$ is then obtained by integrating $\boldsymbol{q}$ with $\boldsymbol{v}^+$. On  (\ref{equ.vnext}), we make the following comments.

The first term $ \frac{1}{h} \boldsymbol{Q}^{-1}\boldsymbol{b}(\boldsymbol{u})$ in (\ref{equ.vnext}) is the predicted velocity by non-contact force $\boldsymbol{b}(\boldsymbol{u})$. The second term is a contact-related term. Recalling  $\dsdf$ in (\ref{equ.lse_dist2}), calculating this term requires comparing the distance between the query  $\boldsymbol{Q}^{-\frac{1}{2}}\boldsymbol{b}(\boldsymbol{u})$ with all facets of the polyhedron constraints in (\ref{equ.polytope}).  Importantly, (\ref{equ.qs_model6}) provides an explicit prediction model, bypassing solving optimization in classic complementarity-based contact models.

Different from   $\csdf_{\mathcal{G}}$ where one can arbitrarily increase the number of supporting planes in object geometry representation $\mathcal{G}$ to control the collision detection accuracy \cite{CVXnet}, The dual cone constraint $\mathcal{C}(\boldsymbol{q})$  in the time-stepping  $\dsdf_{\mathcal{C}(\boldsymbol{q})}$ cannot be arbitrarily altered as it determined by the results of collision detection at system state $\boldsymbol{q}$. {Using the $\dsdf$ may introduce some model approximation error. Thankfully, differentiability of $\dsdf$ makes it easy to use real-world data to adapt the model and minimize sim-to-real discrepancies, as we will present in the next section.}

\vspace{-10pt}
\section{ContactSDF-MPC and \,\, \texorpdfstring{$\dsdf$}{} Learning}\label{sec.planning_and_learning}

\subsection{ContactSDF-MPC}\label{subsection:Complementarity-Free Contact MPC}

With   collision detection $\csdf$  (\ref{equ.lse_dist}), and time stepping prediction $\dsdf$ (\ref{equ.vnext}), we present the ContactSDF-MPC:
\begin{equation}
\begin{aligned}
\min_{\boldsymbol{u}_{0:T\text{-}1}\in[\mathbf{u}_{\text{lb}}, \mathbf{u}_{\text{ub}}]} \quad 
&
\sum\nolimits_{t=0}^{T-1} \textit{c}_t\left ( \boldsymbol{q}_t, \boldsymbol{u}_t \right )+\textit{c}_T\left (\boldsymbol{q}_T\right)
\\ \text{s.t.} \quad &
\boldsymbol{q}_{t+1} = \boldsymbol{q}_t``+"h\boldsymbol{v}_t^+,  \quad \text{given } \boldsymbol{q}_0,
\\ &
\boldsymbol{v}_t^+ \text{ is from } \dsdf_{\mathcal{C}(\boldsymbol{q}_0)} \text{ in } (\ref{equ.vnext}),
\\ & 
\mathcal{C}(\boldsymbol{q}_0) \text{ is from }   \csdf_{\mathcal{G}} \text{ in } (\ref{equ.lse_dist}).
\label{equ.contact_free_mpc}
\end{aligned}
\end{equation}
Note  $\mathcal{C}(\boldsymbol{q}_t)$ is supposed to computed at  predicted $\boldsymbol{q}_t$, $t=0$,...,$T$. However, to reduce the computational complexity,  we assume $\mathcal{C}(\boldsymbol{q}_t)$ is fixed in the relatively short MPC prediction horizon $T$, i.e., we assume $\mathcal{C}(\boldsymbol{q}_t)\approx\mathcal{C}(\boldsymbol{q}_0)$ within MPC prediction \cite{jin2024complementarity}. It should be noted that since both $\csdf$ and $\dsdf$  are explicit and differentiable,  ContactSDF-MPC is capable of handling the time-varying  $\mathcal{C}(\boldsymbol{q}_t)$ in the MPC prediction horizon.

\vspace{-10pt}
\subsection{Learning \texorpdfstring{$\dsdf$}{} Model with On-MPC Data}\label{sec:Learning lse Complementarity-Free Model}

To improve the accuracy of $\dsdf$  in time-stepping prediction,  we propose using the on-MPC data to learn  $\dsdf$  model, based on our prior work \cite{taskdrivenjin}. By learning $\dsdf$   model, we mean learning all model parameters in  $\dsdf$, including 
\begin{equation}
\begin{aligned}
    \boldsymbol{\theta} := \left \{ \boldsymbol{M}_o, \boldsymbol{K}_r, m_o,  \mu, \sigma \right \},
\label{equ:dynamics_parameters_definition}
\end{aligned}
\end{equation}
The learning loss is defined as the prediction loss 
\begin{equation}\label{equ.model_learning_loss}
   \mathcal{L}(\boldsymbol{\theta}, \boldsymbol{\mathcal{D}}_{\text{on-MPC}}) = 
\sum_{(\boldsymbol{q}^{\text{real}}_k, \boldsymbol{u}_k, \boldsymbol{q}^{\text{real}}_{k+1} )\in\boldsymbol{\mathcal{D}}}\left \| \boldsymbol{q}_{k+1}(\boldsymbol{\theta}) - \boldsymbol{q}^{\text{real}}_{k+1}\right \|^2.
\end{equation}
where $\boldsymbol{q}_{k+1}$ is the predicted next state  from  $\dsdf$  given  $(\boldsymbol{q}^{\text{real}}_k, \boldsymbol{u}_k)$. 
Since the  $\dsdf$  is differentiable, the model can be updated using any gradient-based methods.

The pipeline of  $\dsdf$ learning with the on-MPC data follows our prior work \cite{taskdrivenjin}, shown in Fig. \ref{fig:learning_frameware}. It includes two alternative steps: (1) ContactSDF-MPC rollouts on the real system and collects the on-MPC data into a buffer $\boldsymbol{\mathcal{D}}_{\text{on-MPC}}$, and (2)  $\dsdf$  is updated using data in $\boldsymbol{\mathcal{D}}_{\text{on-MPC}}$. 
\vspace{-10pt}
\begin{figure}[htbp]
  \centering
  \includegraphics[scale=0.34]{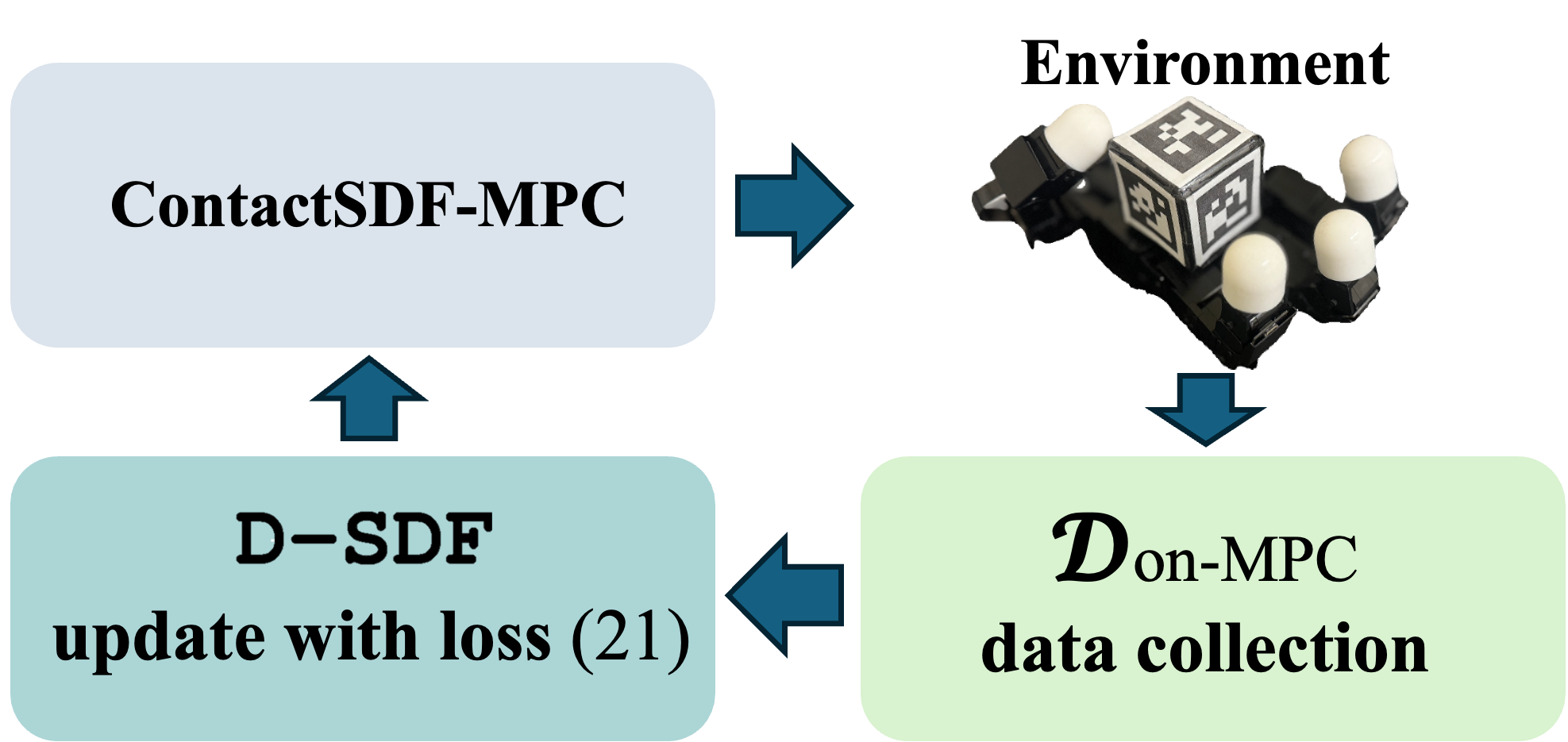}
  \caption{\small Pipeline of  learning $\dsdf$ from on-MPC data \cite{taskdrivenjin}.}
  \label{fig:learning_frameware}
\end{figure}

\vspace{-25pt}
\section{Simulated Experiments}\label{sec.simu_experiments}

We evaluate the ContactSDF-MPC in two contact-rich manipulation environments:  \emph{three-ball manipulation} and \emph{Allegro hand on-palm reorientation}, each with different objects. The environments are built in MuJoCo  \cite{todorov2012mujoco}.  All experiments  run on a PC with an Intel {i9-13900K} chip. The MPC optimization is solved using CasADi \cite{Andersson2019casadi}. The codes are available at \href{https://github.com/asu-iris/ContactSDF}{Link}.

\vspace{-10pt}
\subsection{Three-Ball Manipulation}
\vspace{-10pt}
\subsubsection{Task Setup}\label{sec:ball_setting}

\begin{table}[h]
\begin{center}
\caption{\small {{Target object poses in the three-ball manipulation tasks}}}
\label{table.ball_task_setting}
\begin{threeparttable}
\begin{tabular}{l l l}
    \toprule
     Tasks & Cube/Foambrick & Stick
    \\ 
    \midrule
      {(x,y) position [m]}  
      &  $\{(\pm0.05, \pm0.05)\}$  
      & $\{(\pm0.05, \pm0.05)\}$
      \\
      {z-axis rotation}  [rad]  
      &  $\{{0, \pm\frac{\pi}{4}, \pm\frac{\pi}{2}}\}$  
      & $\{{0, \pm\frac{\pi}{4}, \pm\frac{\pi}{2}}\}$
      \\
      {y-axis flipping}  [rad]
      &  $\{{\pm\frac{\pi}{2}}\}$  
      & $\{\frac{3}{4}\pi, \pi\}$
      \\
      \bottomrule
\end{tabular}
\end{threeparttable}
\end{center}
\end{table}

  \vspace{-30pt}
\begin{figure}[htbp]
  \centering
  \includegraphics[scale=0.3]{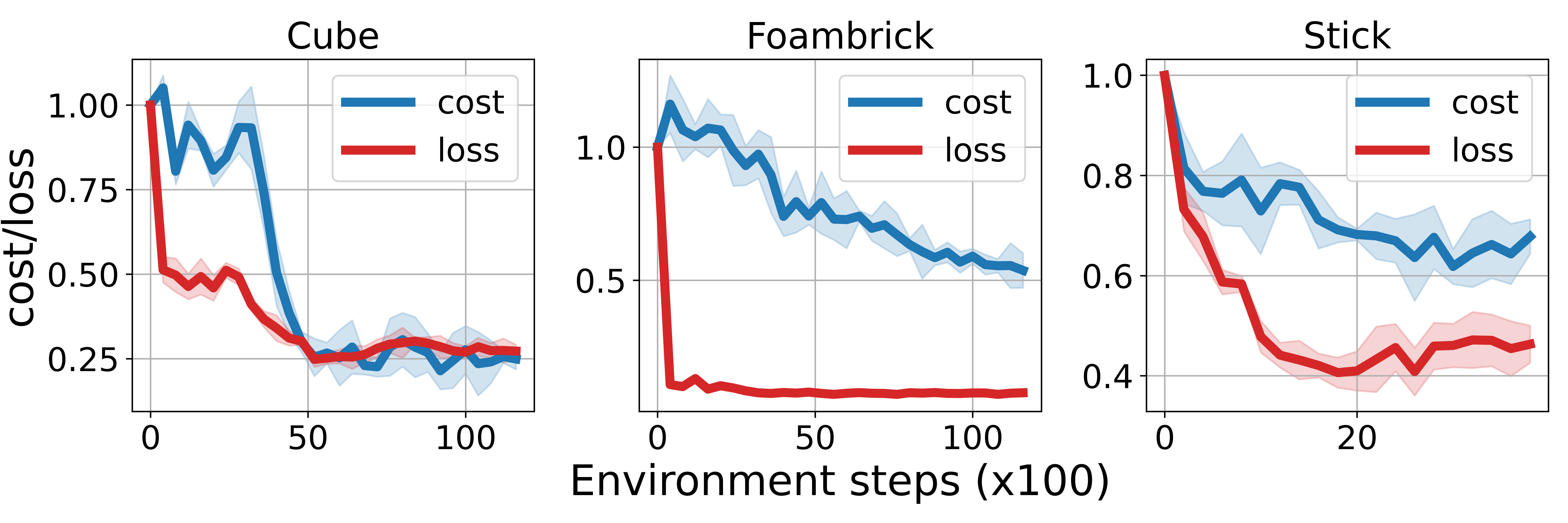}
  \caption{\small Learning $\dsdf$ with on-MPC data in the three-ball manipulation tasks. Left and right panel shows the normalized accumulated cost of environment rollout, evaluated by $c_T$  (\ref{equ:mpc_detail_cost_fn}), and normalized model prediction loss (\ref{equ.model_learning_loss}), respectively. Each plot is based on five learning trials.} 
  \label{fig:ball_learning_data_plot}
\end{figure}

\begin{table*}[h]
\begin{center}
\caption{Comparison results for the three-ball manipulation tasks}
\label{table.ball_comparison_results}
\begin{threeparttable}
\begin{tabular}{l l l l l l l}
    \toprule
    \multirow{3}{*}{Case} & 
    \multicolumn{2}{c}{\textbf{Terminal position error [mm]$\downarrow$ }} & 
    \multicolumn{2}{c}{\textbf{Terminal orientation error [rad]$\downarrow$ }} & 
    \multicolumn{2}{c}{\textbf{MPC solving cost [ms] $\downarrow$ }} \\
    \cmidrule(rl){2-3} \cmidrule(rl){4-5} \cmidrule(rl){6-7} 
     &
     {ContactSDF-MPC}&
     {QPModel-MPC}&
     {ContactSDF-MPC}&
     {QPModel-MPC}&
     {ContactSDF-MPC}&
     {QPModel-MPC} \\
    \midrule
      Cube     & \textbf{13.62}$\pm$\textbf{3.14}  & {26.11}$\pm${28.24}             
      & \textbf{0.11}$\pm$\textbf{0.11}  & {0.24}$\pm${0.35} 
      & \textbf{31.77}$\pm$\textbf{9.13}  & {80.76}$\pm${14.19}
      \\
      Foambrick   & {8.93}$\pm${2.31}  & \textbf{8.59}$\pm$\textbf{3.17}
      & 0.24$\pm$0.02  & \textbf{0.19}$\pm$\textbf{0.05} 
      & \textbf{24.95}$\pm$\textbf{8.00}  & 51.64$\pm$15.81  
      \\
      Stick    & \textbf{22.06}$\pm$\textbf{5.90}  & 30.54$\pm$8.72 
      & \textbf{0.26}$\pm$\textbf{0.19}  & 0.37$\pm$0.22 
      & \textbf{52.81}$\pm$\textbf{5.99}  & 91.94$\pm$10.56   
      \\
      \bottomrule
\end{tabular}
\begin{tablenotes}
\centering
\item[] The results of each object are reported using 7 random trials
including 5 turning and 2 flipping targets. The position and orientation errors are computed at the last step of MPC rollout.  
The relaxation factor in {QPModel-MPC} is set $\epsilon{=}10^{-4}$ for  best performance.
\end{tablenotes}
\end{threeparttable}
\end{center}
\vspace{-10pt}
\end{table*}

As in Fig. \ref{fig:ball_finished_tasks}, the three-ball manipulation involves three balls (red, green,  blue), each with 3 DoFs actuated by a low-level position controller with gravity compensation, and an object on ground. Three objects are used:   cube,  foambrick, and stick, with their geometries given. The task goals are that three balls translate, rotate, and/or flip an object from an initial pose to a random on-ground target (shown in transparency in Fig. \ref{fig:ball_finished_tasks}) in Table. \ref{table.ball_task_setting}, where each target  pose is a combination of a target position and rotation.


\subsubsection{ContactSDF setting}\label{sec.mpc_cost_definition}

The system state $\boldsymbol{q}$ involves object pose  $\boldsymbol{q}_o \in \mathbb{R}^7$ and the positions of three balls $\boldsymbol{q}_r \in \mathbb{R}^9$; the control input $\boldsymbol{u} \in \mathbb{R}^9$ is the desired position displacement of three balls, which is sent to the low-level position controller.  
The time interval in $\dsdf$ is set to $h = 0.1$s. Other parameters (\ref{equ:dynamics_parameters_definition}) in $\dsdf$ will be learned using on-MPC data.


In  $\csdf$, for contacts between the object and three balls, the balls are used as query points $\boldsymbol{x}_{\text{query}}$. For  contact between the object and ground, the query points are the ground points, which are evenly sampled from the object’s  projection area on the ground.
For each object, its  $\mathcal{G}$ in (\ref{equ.dist_convexset}) is  generated from its mesh. 
{The frictional cone in (\ref{equ:qs_dyn_simplified}) is linearized with $n_d=4$.}

In ContactSDF-MPC (\ref{equ.contact_free_mpc}),   $T{=}4$,   $\mathbf{u}_{\text{lb}}{=-}0.01$ and $\mathbf{u}_{\text{ub}}{=}0.01$.  The  path and final costs  are defined \cite{jin2024complementarity} as
\begin{equation}
\begin{aligned}
\textit{c}_t(\boldsymbol{q},\boldsymbol{u}) = &
\omega_c
c_{\text{contact}} + 
\omega_gc_{\text{grasp}} +
\omega_u \norm{\boldsymbol{u}}^2
\\
\textit{c}_T(\boldsymbol{q}) = & 
\omega_p\norm{\mathbf{p}_{{o}}-\mathbf{p}_{{o}}^{\text{target}}}^2 + 
\omega_q
\left ( 1 - (\mathbf{q}_o\tran \mathbf{q}^{\text{target}})^2 
\right ).
\label{equ:mpc_detail_cost_fn}
\end{aligned}
\end{equation}
respectively.  The path cost $\textit{c}_t(\boldsymbol{q},\boldsymbol{u})$ has three terms: (i) The contact cost term, 
$c_{\text{contact}} \text{=}\sum\nolimits_{i=0}^{2} 
\norm{\mathbf{p}_{\text{ball},i} {-} \mathbf{p}_{{o}}}^2$, is to encourage the contact between the three balls and object. (ii) The grasp cost term \cite{jin2024complementarity}, defined as $   c_{\text{grasp}}\text{=}\norm{{\mathbf{ d}}_0{+} {\mathbf{ d}}_1{+} {\mathbf{ d}}_2 }^2$,
is to encourages a grasp closure; here, $\mathbf{d}_i{=}\boldsymbol{R}_o\tran\left(\mathbf{q}_{\text{ball},i} {-} \mathbf{p}_o\right)/\norm{\left(\mathbf{q}_{\text{ball},i} {-} \mathbf{p}_o\right)}$ is a  unit directional vector from object $\mathbf{p}_o$ to each ball position $\mathbf{q}_{\text{ball},i}$. (iii) $\norm{\boldsymbol{u}}^2$ is the control effort.  The final cost $\textit{c}_T(\boldsymbol{q})$ defines the distance to the target $(\mathbf{p}_o^{\text{target}},\mathbf{q}_o^{\text{target}})$. The cost weights $\{ \omega_c, \omega_g, \omega_u, \omega_p, \omega_q \}$ are set  based on physical scale: $\{ 1, 0.1, 1, 10000, 1000 \}$ for cube, $\{ 1, 0.1, 1, 10000, 5000 \}$ for  foambrick, and $\{ 1, 0.1, 1, 500, 100 \}$ for stick.

\smallskip
\subsubsection{Results and Comparison}
Following the procedure in Fig. \ref{fig:learning_frameware}, we first learn the $\dsdf$ model (i.e., parameter $\boldsymbol{\theta}$ in (\ref{equ:dynamics_parameters_definition})) with on-MPC data for each object manipulation. Here, we set the ContactSDF-MPC rollout length $H=100$.
 $\dsdf$  is updated every  4  on-MPC rollouts (i.e., every $400$ environment steps). 
Each rollout is tasked with a target (position + rotation) uniformly sampled from Table \ref{table.ball_task_setting}. We show the learning results in Fig. \ref{fig:ball_learning_data_plot}. The plots  show that a $\dsdf$  can be successfully learned with less than  5k  environment steps. With the learned $\dsdf$ model, we next evaluate  ContactSDF-MPC.  

\begin{figure}[htbp]
  \centering
  \includegraphics[scale=0.3]{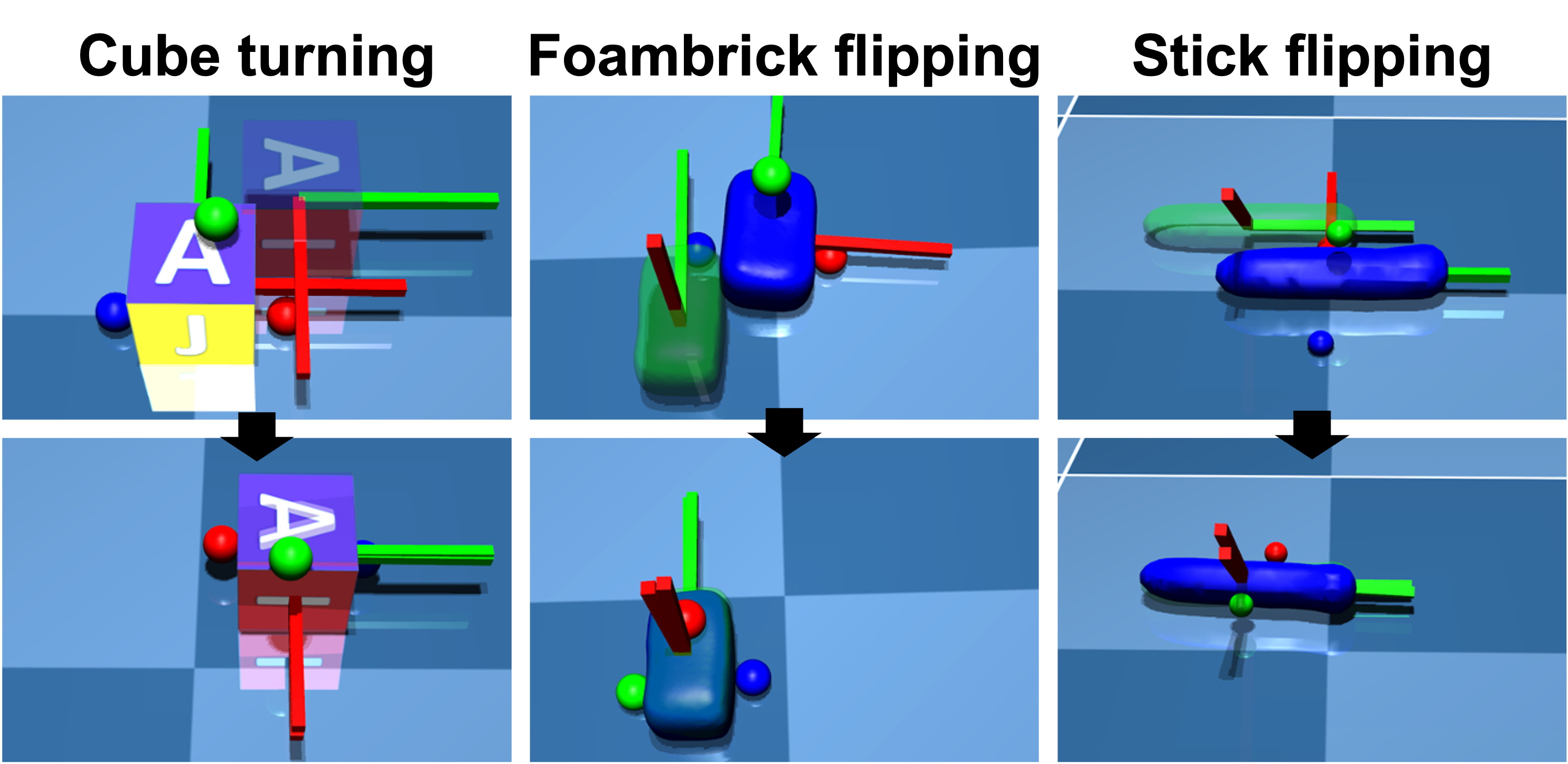}
  \caption{\small Evaluation examples for ContactSDF-MPC in the three-ball manipulation.   The first and second rows show the initial and final scenes of a task, respectively. 
  }
  \label{fig:ball_finished_tasks}
  \vspace{-5pt}
\end{figure}

{In evaluation, we compare ContactSDF-MPC with the MPC that uses the QP-based model (\ref{equ:qs_dyn_simplified}). We refer to the latter as {QPModel-MPC}. To solve QPModel-MPC with existing solver \cite{Andersson2019casadi}, we replace the QP model with its KKT conditions with a relaxation $\epsilon$ for complementarity conditions similar to \cite{fast_mpc}. This is equivalent to using a barrier-smoothed quasi-dynamics model with fixed barrier parameter. The relaxation parameter is set to $\epsilon=10^{-4}$ to strike a balance between the accuracy and computational efficient.} The ContactSDF-MPC and {QPModel-MPC} share  the same cost functions ($\ref{equ:mpc_detail_cost_fn}$) and    $\boldsymbol{\theta}$ learned. 
For each object, we run both MPCs for 7  trials, each with a  random target uniformly sampled from Table \ref{table.ball_task_setting}, and  rollout length  $H=200$ for cube and $H=300$ for foambrick and stick. The  results are  in Table \ref{table.ball_comparison_results}. Fig. \ref{fig:ball_finished_tasks} visualizes some evaluation examples. 

{The results in Table \ref{table.ball_comparison_results} and Fig. \ref{fig:ball_finished_tasks} demonstrate ContactSDF-MPC achieves noticeably higher manipulation accuracy compared to QPModel-MPC. Specifically, ContactSDF-MPC reduces average position error by 31.6\% and orientation error by 24.9\% across objects. Although in the foambrick case, ContactSDF MPC is outperformed slightly 
by QPModel-MPC (position < 0.5 [mm], rotation < 0.05 [rad]). We attribute this to the approximation error of ContactSDF which sacrifices modeling accuracy comparing to QP-based model. Moreover, ContactSDF-MPC has a substantial optimization speed advantage,  103\% faster on average than QPModel-MPC. This speed advantage is primarily due to the elimination of complementarity in ContactSDF.}

\vspace{-20pt}
\subsection{Allegro Hand On-palm Reorientation}

\vspace{-15pt}
\begin{figure}[ht]
  \centering
  \includegraphics[scale=0.28]{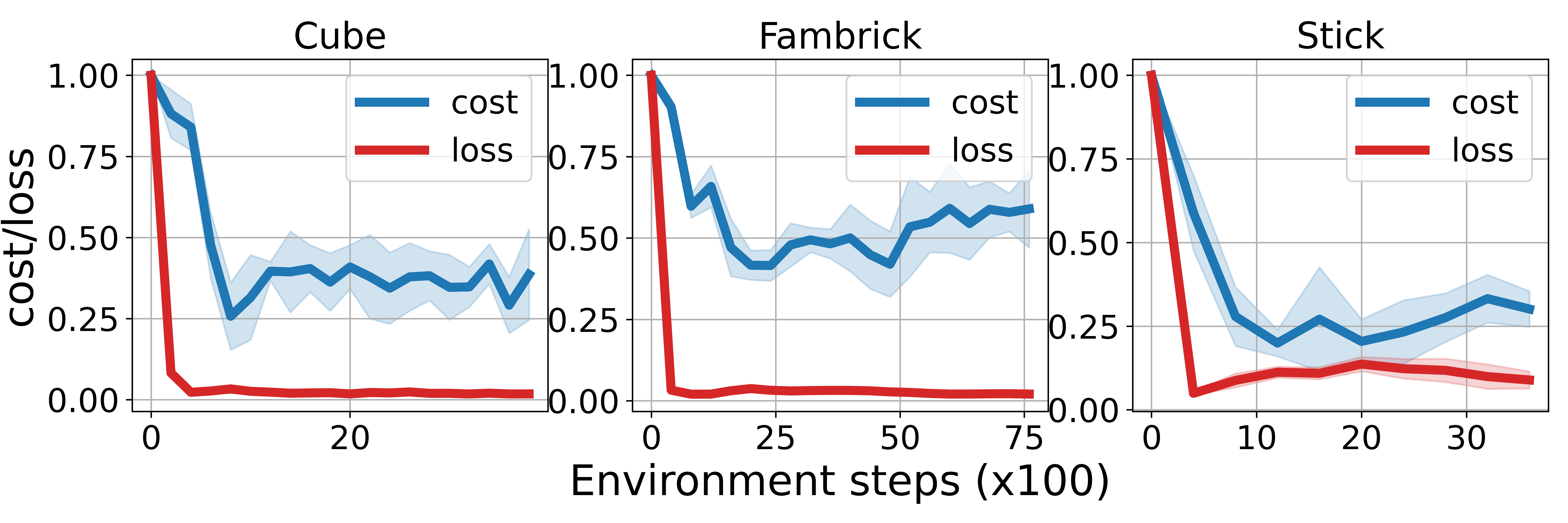}
  \caption{\small Learning $\dsdf$ with on-MPC data in the Allegro hand on-palm reorientation tasks.  The blue and red lines show the normalized accumulated cost of environment rollout, evaluated by $c_T$  (\ref{equ:mpc_detail_cost_fn}), and normalized model prediction loss (\ref{equ.model_learning_loss}) along with the learning steps. Each plot is based on five learning trials, showing the mean and standard deviation.}
  \label{fig:hand_learning_data_plot}
  \vspace{-10pt}
\end{figure}

\begin{figure}[ht]
  \centering
  \includegraphics[scale=0.07]{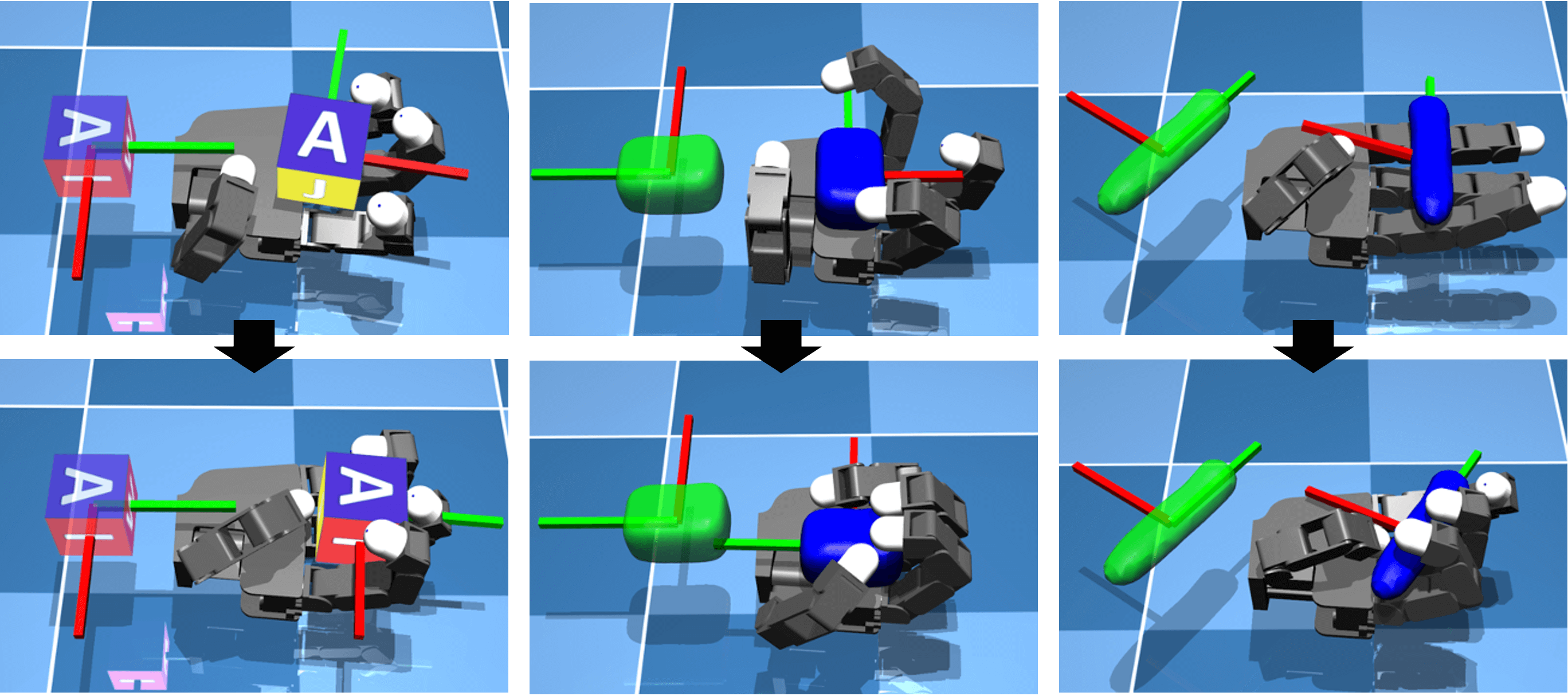}
  \caption{\small Evaluation examples of ContactSDF-MPC for Allegro hand manipulation of different objects.  The first and the second rows show the initial and final scenes of each task, respectively.}
  \label{fig:hand_finished_tasks}
\end{figure}

\subsubsection{Task Setup}
As in Fig. \ref{fig:hand_finished_tasks}, the Allegro robotic hand has four 4-DoF fingers with the palm facing upward. We consider the on-palm object reorientation task with three objects used in the three-ball manipulation. The task involves reorienting objects from an initial orientation on the palm to given target orientations.  We specifically consider reorientation around z-axis (yaw rotation), with the target given in Table. \ref{table.hand_task_setting}.

\vspace{-10pt}
\begin{table}[h]
\begin{center}
\caption{\small {Targets in Allegro hand on-palm reorientation tasks}}
\label{table.hand_task_setting}
\begin{threeparttable}
\begin{tabular}{l l l}
    \toprule
     Target & Cube/Foambrick & Stick
    \\ 
    \midrule
      {z-axis (yaw) rotation}
      &  $\{{\pm\frac{\pi}{4}, \pm\frac{\pi}{3}, \pm\frac{\pi}{2}}\}$  
      & $\{\pm0.2\pi, \pm0.25\pi, \pm0.3\pi \}$
      \\
      \bottomrule
\end{tabular}
\end{threeparttable}
\end{center}
\vspace{-15pt}
\end{table}

\subsubsection{ContactSDF Setting}
The system state $\boldsymbol{q}\in\mathbb{R}^{23}$, including 
the object pose $\boldsymbol{q}_o \in \mathbb{R}^7$ and robot joint position $\boldsymbol{q}_r \in \mathbb{R}^{16}$. $\boldsymbol{u} \in \mathbb{R}^{16}$ is the desired joint displacement,  sent to a low-level joint PD controller. The   parameters  $\boldsymbol{\theta}$ (\ref{equ:dynamics_parameters_definition}) in $\dsdf$ will be learned using on-MPC policy data. {In  $\csdf$,
the contact query points between the hand and the object spread across the surfaces of every fingers and palm surface. The supporting plane number $I$ for three objects are 12/50/500 and number of contact query points is 525 in total, which lead to contact query time 0.4/0.9/6.2[ms].}
Filtering are applied to remove distant queries. 


The setting of the {ContactSDF-MPC} (\ref{equ.contact_free_mpc}) follows the three-ball manipulations. The cost functions are the same as (\ref{equ:mpc_detail_cost_fn}), except that the grasp cost term is removed as it is not critical for on-palm orientation tasks. The cost weights $\{ \omega_c, \omega_u, \omega_p, \omega_q \}$ are  $\{ 0.5, 0.1, 1000, 10000 \}$ for cube, $\{ 1, 0.2, 500, 5000 \}$ for foambrick and $\{ 0.5, 0.2, 1000, 10000 \}$ for stick.

\subsubsection{Results}
Similar to the procedure in the previous three-ball manipulation, all  parameters $\boldsymbol{\theta
}$ in the $\dsdf$ are learned for each object task. The MPC rollout length is set to $H=100$. In each rollout, the reorientation target is uniformly sampled from  Table. \ref{table.hand_task_setting}. The learning results are shown in Fig. \ref{fig:hand_learning_data_plot}. 

\vspace{-10pt}
\begin{table}[h]
\begin{center}
\caption{\small {Evaluation performance for Allegro In-hand reorientation}}
\label{table.in_hand_simu_performance}
\begin{threeparttable}
\begin{tabular}{l l l l}
    \toprule
     Metrics & Cube & Foambrick & Stick
    \\ 
    \midrule
      {Terminal orientation err. [rad]}  &   0.08$\pm$0.06 & 0.17$\pm$0.15 & 0.28$\pm$0.21
      \\
      {MPC running frequency [Hz]}  &   40.6$\pm$17.1 & 40.3$\pm$12.9 & 46.9$\pm$15.3
      \\
      \bottomrule
\end{tabular}
\end{threeparttable}
\end{center}
\vspace{-10pt}
\end{table}

With the learned $\dsdf$, we evaluate the performance of {ContactSDF-MPC}. For each object,  ContactSDF-MPC is evaluated for 6 trials, each with a random target uniformly sampled from Table. \ref{table.hand_task_setting}, rollout length $H=150$ for cube and $H=200$ for foambrick and stick. The evaluation results are in Table. \ref{table.in_hand_simu_performance}. Fig. \ref{fig:hand_finished_tasks} visualizes some evaluation examples.

With the learning and evaluation results given above, we draw the following conclusions:
(i)   Fig. \ref{fig:hand_learning_data_plot} shows the ContactSDF model converges quickly within about 2k environment steps. 
{(ii) Table. \ref{table.in_hand_simu_performance} illustrates ContactSDF-MPC achieves high manipulation accuracy for the Allegro on-palm reorientation tasks, although the geometry that directly constructed from meshes may introduce contact noisy: the average reorientation error is 0.08 rad for the cube, 0.17 rad for foambrick. For the stick, despite non-equilibrium targets, it still accomplishes the rotating task with an averaged angle error of 0.28 rad.} 
(iii) 
Table \ref{table.in_hand_simu_performance} shows that ContactSDF-MPC achieves a real-time speed for MPC of Allegro in-hand manipulation, and the average MPC speed is above 40 Hz.

\vspace{-10pt}
\section{Hardware Experiments}\label{sec.hardware_experiments}

\subsection{Hardware  Implementation}
\begin{figure}[t]
  \centering
  \includegraphics[scale=0.07]{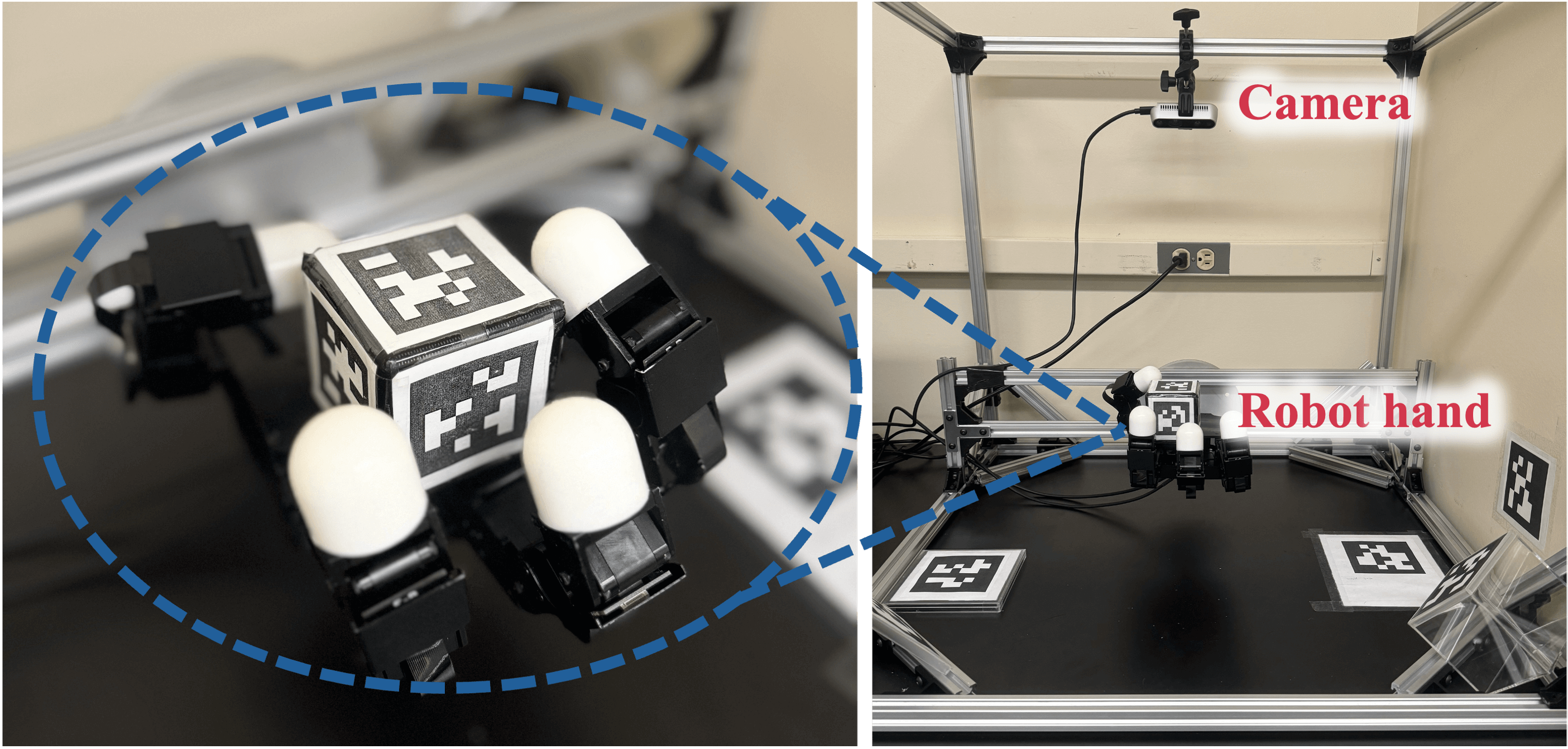}
  \caption{\small Allegro hand (V4) hardware system.}
  \label{fig:hardware_setup}
\end{figure}

\begin{figure}[t]
  \centering
  \includegraphics[scale=0.30]{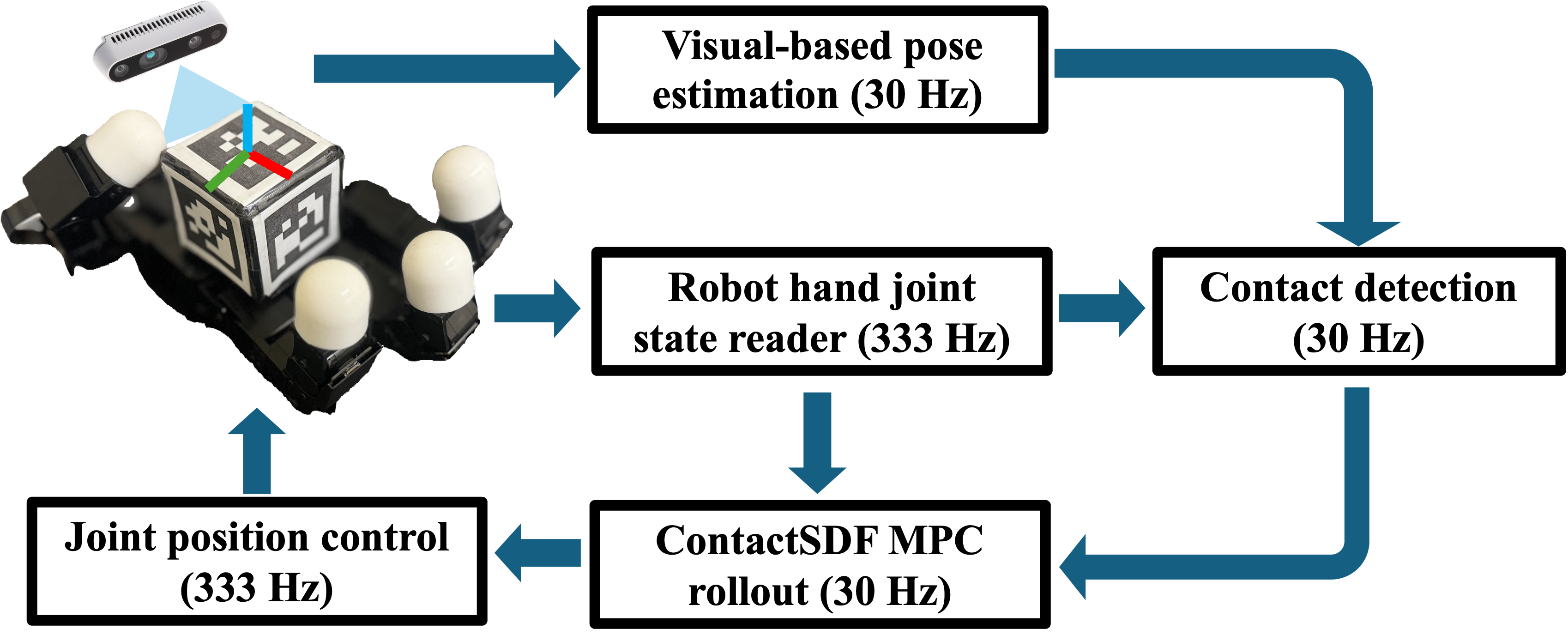}
  \caption{\small System diagram for the hardware implementation. Different blocks indicate separate processes connected via arrows that indicate communication via ROS.}
  \vspace{-10pt}
  \label{fig:framework}
\end{figure}

The hardware Allegro hand system is shown in Fig. \ref{fig:hardware_setup}, and each 4-DoF finger is actuated by a low-level joint position PD controller running at  333Hz.  An RGBD camera (RealSense D435i )  is used for the object pose estimation. We use a cube object of ($56\times56\times56 \text{mm}$), with  faces attached with  AprilTags \cite{olson2011apriltags} for  pose  estimation with TagSLAM \cite{tagslam}. The task involves reorienting the cube in $z$-axis (yaw angle). The system  modules are shown in Fig. \ref{fig:framework}. The object pose estimation has a frequency of 30 Hz.  We synchronize the object pose estimation with  $\csdf$ contact detection as the collision detection is dependent of object pose. ROS is used for communication.

The ContactSDF setting follows that in the previous simulation. ContactSDF-MPC (\ref{equ.contact_free_mpc}) setting is in Table \ref{table.real_hand_manipulation_other_params}.

\vspace{-10pt}
\begin{table}[ht]
\begin{center}
\caption{\small{Parameters of hardware Allegro hand experiments}}
\label{table.real_hand_manipulation_other_params}
\begin{threeparttable}
\begin{tabular}{l l}
    \toprule
    \textbf{Parameter}  & Value
    \\ 
    \midrule
      Time interval $h$ in $\dsdf$           & 0.1 [s]   
      \\
            MPC prediction horizon  $T$         & 3   
      \\
                  Cost weights (\ref{equ:mpc_detail_cost_fn})   $\{ \omega_c, \omega_u, \omega_p, \omega_q \}$         & $ \{1, 0.2, 500, 5000\}$ 
      \\
          ContactSDF-MPC control bounds & $\boldsymbol{u}_{\text{lb}}{=-}0.15$  \text{ and } $\boldsymbol{u}_{\text{ub}}{=}0.15$   
      \\
            Incremental sub-goal setting     & current yaw angle $\pm90^\circ$
      \\
      ContactSDF-MPC rollout  length $H$      & 100  
        \\
      $\boldsymbol{\mathcal{D}}_{\text{buffer}}$ buffer size  & 400  [environment steps]
      \\
      \bottomrule
\end{tabular}
\end{threeparttable}
\end{center}
  \vspace{-10pt}
\end{table}

\subsection{Learning \texorpdfstring{$\dsdf$}{} on Hardware on-MPC Data}
Unlike in simulation where one can conveniently reset the cube state during the learning,  resetting the cube state on hardware has to be manual. To reduce  manual cube resetting, our task goal in the learning stage is to continuously rotate the cube around the z-axis. We achieve this by incrementally setting sub-goals, each involving adding $90^{\circ}$ (CCW) to the cube's yaw angle at the end of the a MPC rollout.  When a rollout ends, the fingers are also reset to their initial positions. 

We learn  $\dsdf$  \emph{from scratch} using hardware on-MPC data. 
The $\dsdf$  is updated every four hardware MPC rollouts. Each rollout corresponds to an incremental sub-goal and has a length of 100 steps.
{The learning results are shown in Fig. \ref{fig:Hardware_Learning}. The plots in Fig. \ref{fig:Hardware_Learning} show that the ContactSDF model is efficiently learned with around 2k hardware steps less than 2 minutes}. 

\vspace{-10pt}
\begin{figure}[htbp]
  \centering
  \includegraphics[scale=0.21]{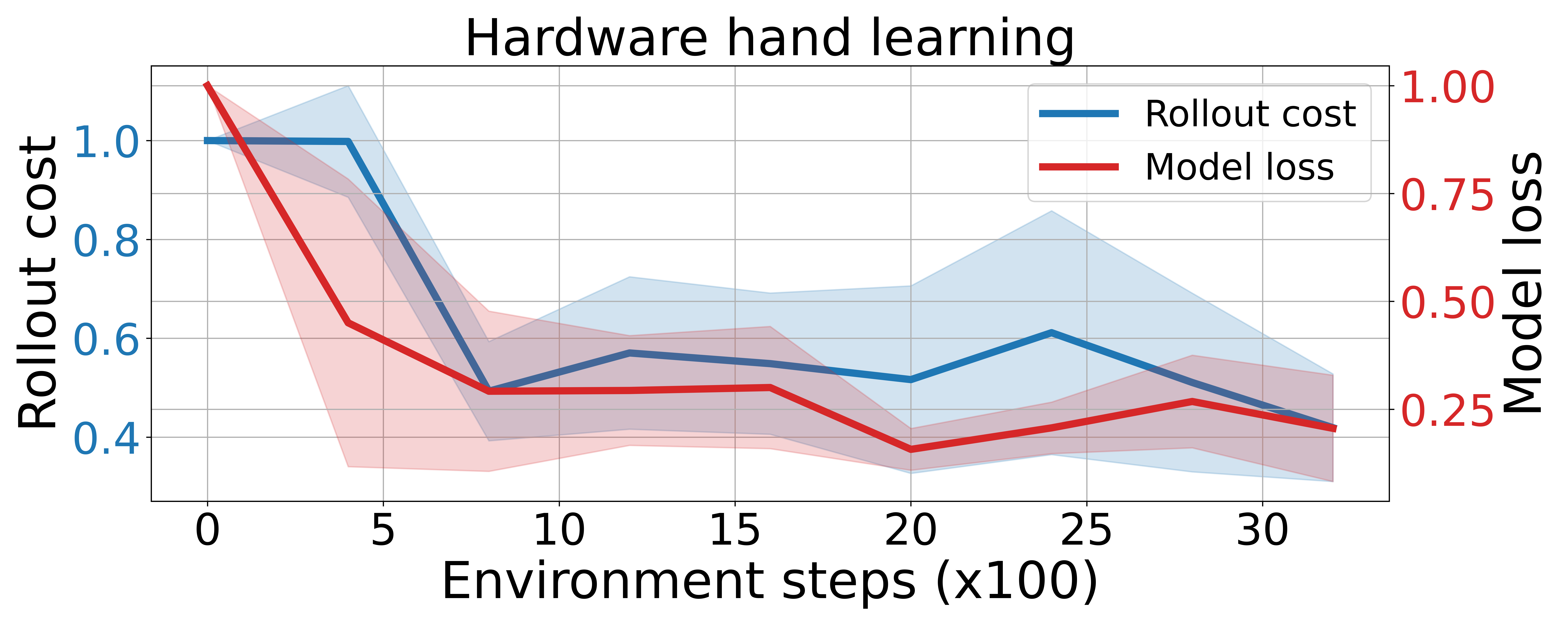}
  \caption{\small Learning $\dsdf$ from hardware data of Allegro hand continuous reorientation. The  
 accumulated cost (evaluated by $c_T$)  and normalized model  loss are both normalized.}
  \label{fig:Hardware_Learning}
\end{figure}


\vspace{-20pt}
\begin{figure}[ht]
  \centering
  \includegraphics[scale=0.33]{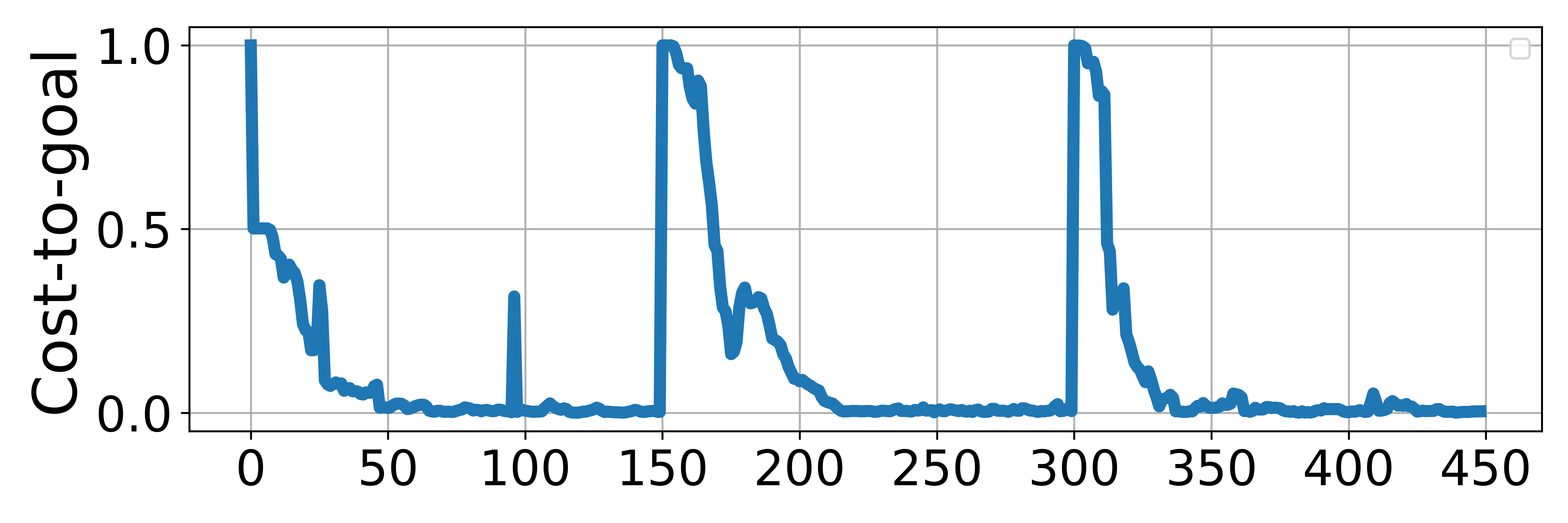}
  \caption{\small An example of continuous reorientation task as Fig. \ref{quadruped_hardware} right, evaluated by the normalized cost-to-goal at each rollout steps.}
  \label{fig:continuous_demo_plot}
  \vspace{-10pt}
\end{figure}

\vspace{-10pt}
\subsection{Evaluation of  ContactSDF-MPC for Single Rotations}
With the learned $\dsdf$, we evaluate the performance of ContactSDF-MPC for reorientation tasks given single $90^\circ$ (CCW) or ${-}90^\circ$ (CW) rotation goals. For each goal, we perform 5 trials. In each trial, the cube is randomly placed around the palm center, and we run ContactSDF-MPC for one rollout   of length $H=100$. The evaluation results are  in Table \ref{table.hardware_performance}. 

\begin{table}[h]
\begin{center}
\caption{\small {Evaluation results for  two orientation goals}}
\label{table.hardware_performance}
\begin{threeparttable}
\begin{tabular}{l l l}
    \toprule
      & CCW & CW 
    \\ 
    \midrule
      {Terminal orientation err. [rad]}  &   0.11$\pm$0.05 & 0.19$\pm$0.15 
      \\
      {MPC running frequency [Hz]}  &   47.2$\pm$12.8 & 49.5$\pm$11.3
      \\
      \bottomrule
\end{tabular}
\end{threeparttable}
\vspace{-10pt}
\end{center}
\end{table}

Table. \ref{table.hardware_performance} shows a high reorientation accuracy of ContactSDF-MPC for single rotation tasks. The average reorientation error is 0.15 rad for both rotation goals. Importantly,  the ContactSDF-MPC can run 50Hz on the hardware. 

\vspace{-10pt}
\subsection{Evaluation of  ContactSDF-MPC for Continuous Rotation}
Next, we evaluate  ContactSDF-MPC for \emph{continuous} rotation of the cube without human resetting. Similar to  learning stage, the sub-goal is incremented by adding $90^{\circ}$ to the cube yaw angle at the end of each MPC rollout ($H=150$). Also when a MPC rollout ends, the fingers are reset to their initial  positions. Initially the cube is randomly placed on the palm.

The results are shown in the right panel of Fig. \ref{quadruped_hardware}  and in Fig. \ref{fig:continuous_demo_plot}. In Fig. \ref{quadruped_hardware} right, the first row depicts the snapshots of the hardware system at different rollout steps, and the second row shows the  incremental rotation target (black line) and actual cube yaw angle (blue line). As plotted, there are three incremental sub-goals; after each sub-goal is achieved at the end of an MPC rollout, the target cube angle is updated to the next by adding $90^\circ$. The first row in Fig. \ref{fig:continuous_demo_plot} reports the normalized cost-to-goal of the system, evaluated using final cost $c_{T}(\cdot)$ in (\ref{equ:mpc_detail_cost_fn}) at every rollout step. 

\vspace{-10pt}
\section{Conclusion and Future Work}\label{sec.conclusion}

We proposed the ContactSDF, a differentiable contact model for efficient learning and control of contact-rich manipulation systems. By using SDFs to approximate both collision detection and time-stepping prediction, ContactSDF eliminates the hybrid and non-smooth routines in classic complementarity-based contact formulation. ContactSDF can be readily integrated with existing learning and  control methods for efficiently solving dexterous manipulation. The efficiency of ContactSDF  has been extensively demonstrated in both simulation and hardware experiments. Several areas of the ContactSDF remain unexplored, including learning object geometry SDF and generalizing the hardware experiments to more complex objects. We will explore those aspects in the future work.




%


\vspace{-10pt}
\bibliographystyle{unsrt}
\bibliography{mybib} 

%




\end{document}